\documentclass{article} 
\usepackage{iclr2023_conference,times}

\usepackage{amsmath,amsfonts,bm}

\def\eqref#1{equation~\ref{#1}}
\def\Eqref#1{Equation~\ref{#1}}

\def\1{\bm{1}}

\DeclareMathAlphabet{\mathsfit}{\encodingdefault}{\sfdefault}{m}{sl}
\SetMathAlphabet{\mathsfit}{bold}{\encodingdefault}{\sfdefault}{bx}{n}

\usepackage{hyperref}
\usepackage{url}
\usepackage{caption}
\usepackage{subcaption}
\usepackage{graphicx}
\usepackage{xcolor}
\usepackage{arydshln}
\usepackage{amsmath}
\usepackage{booktabs}

\newcommand{\method}[1]{\textit{LADS}}

\definecolor{mycolor}{RGB}{229, 83, 0}

\newcommand{\dtrain}{D_\text{training}}
\newcommand{\dnew}{D_\text{unseen}}
\newcommand{\ttrain}{t_\text{training}} 
\newcommand{\tnew}{t_\text{unseen}}
\newcommand{\ctext}{\text{t}}

\newcommand{\clipimg}[1]{\text{CLIP}_\text{img}(#1)}
\newcommand{\img}[1]{I_{\theta}(#1)}
\newcommand{\cliptext}[1]{\text{CLIP}_\text{text}(#1)}
\newcommand{\txt}[1]{T_{\theta}(#1)}

\newcommand{\faug}{f_\text{aug}}

\title{Using Language to Extend to Unseen Domains}

\author{Lisa Dunlap, Clara Mohri \thanks{equal contribution} \\
UC Berkeley \\
\texttt{\{lisabdunlap,cmohri\}@berkeley.edu} \\
\And 
Aditi Raghunathan \\
Carnegie Mellon University \\
\texttt{raditi@cmu.edu} \\
\AND
Han Zhang, Devin Guillory, Trevor Darrell, Joseph E. Gonzalez, Anna Rohrbach \\
UC Berkeley \\
\texttt{\{pariszhang,dguillory,trevordarrell,jegonzal,anna.rohrbach\}@berkeley.edu} \\
}

\iclrfinalcopy 
\begin{document}
\maketitle

\begin{abstract}
It is expensive to collect training data for every possible domain that a vision model may encounter when deployed. 
We instead consider how simply \emph{verbalizing} the training domain (e.g. ``photos of birds'') as well as domains we want to extend to but do not have data for (e.g. ``paintings of birds'') can improve robustness. 
Using a multimodal model with a joint image and language embedding space, our method \method{} learns a transformation of the image embeddings from the training domain to each unseen test domain, while preserving task relevant information.
Without using any images from the unseen test domain, we show that over the \emph{extended} domain containing both training and unseen test domains, \method{} outperforms standard fine-tuning and ensemble approaches over a suite of four benchmarks targeting domain adaptation and dataset bias. Code is available at \url{https://github.com/lisadunlap/LADS}. 


\end{abstract}

\begin{figure}[ht]
    \centering
    \includegraphics[width=0.85\textwidth]{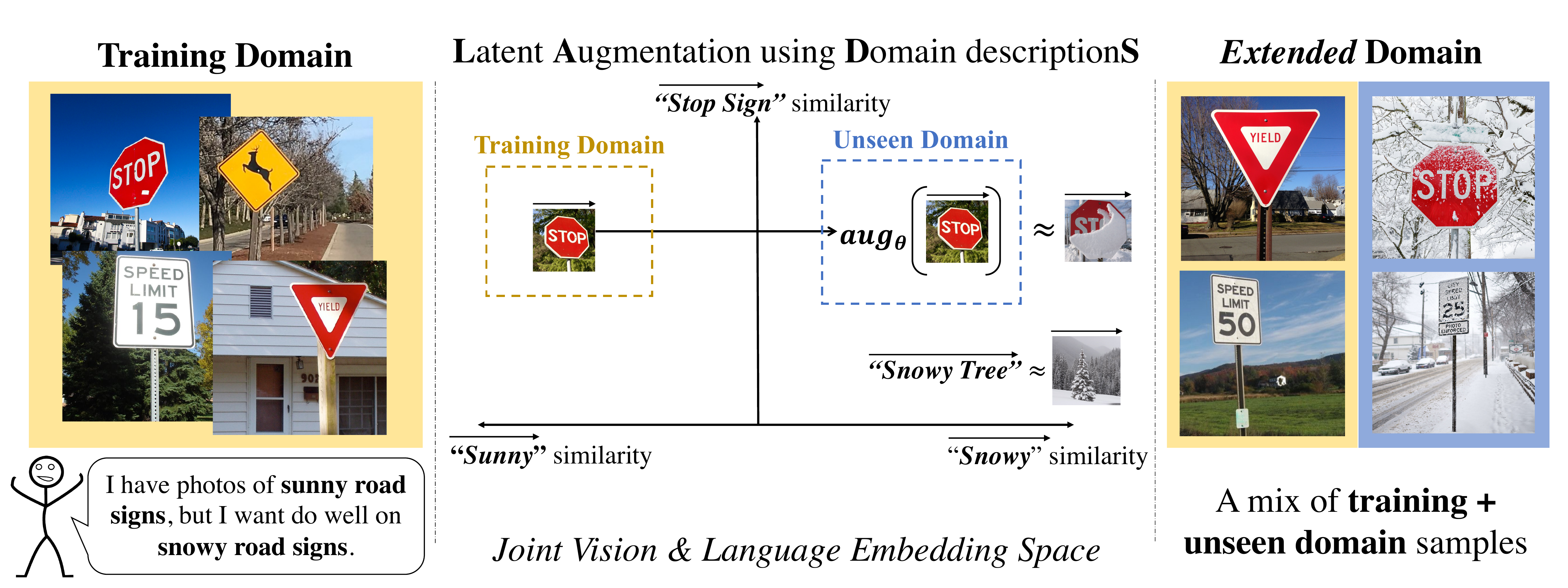}
    \caption{Consider a model trained to recognize road signs in sunny weather. We aim to \emph{extend} to a new domain of snowy weather. Our method \method{} (Latent Augmentation using Domain descriptionS) leverages a multimodal model's knowledge of the classes and the domain shift verbalized in natural language (``sunny'' to ``snowy'') to train an augmentation network without any samples from the unseen test domain. This network is used to translate multimodal \emph{image embeddings} from the training domain to the unseen test domain, while retaining class-relevant information. Then, real and augmented embeddings are used jointly to train a classifier.}
    \label{fig:teaser}
\end{figure}

\section{Introduction}
\label{sec:intro}
The ability to extend a model beyond the domain of the training data is central to building robust computer vision models.
Methods for dealing with unseen test distributions often require leveraging additional image data, but linguistic knowledge of the anticipated domain shift is  much cheaper and easier to obtain. 
For example, in many settings, the training images are collected in certain conditions (e.g., daylight, clear weather, ...) but our sensors may also experience less common but easy to anticipate conditions (e.g., night, snow, haze, illustrations, ...).
Directly collecting or creating data in all possible anticipated settings is often prohibitively expensive.
Thus it is of great interest how one can \textit{linguistically extend} to unseen domains: that is, to utilize language to improve performance on an unseen test domain without sacrificing performance on the training domain.

The use of language in domain generalization has generated significant interest with the development of large vision-language models such as CLIP~\citep{Radford2021LearningTV}, Flamingo~\citep{flamingo}, and ALIGN~\citep{align}, which allow users to create zero-shot classifiers using only class names.
However, while these models have been shown to achieve remarkable cross-domain generalization, their zero-shot classifiers often perform far worse than models trained for a particular downstream task~\citep{Radford2021LearningTV, kumar2022finetuning}.
When training data is available for the downstream task, a common practice is to fine-tune these models on the training data.
While this significantly improves in-domain accuracy, it degrades performance on unseen domains.

We show that it is possible to leverage the domain-level knowledge 
(e.g. sunny environments vs. snowy environments in our example)
contained in CLIP or similar models to deal with a variety of domain shifts in a way that requires no data from the new test domain, exploits the labeled training data, and is fast to train. 
Our method only requires users to input text descriptions of the training and unseen test domains (e.g. ``a sunny stop sign'' and ``a snowy stop sign'') along with their training data. 
To achieve language-guided domain generalization, we leverage the broad domain knowledge encoded in CLIP coupled with its shared image-language embedding space to perform \emph{latent feature augmentation} of the training set.

More precisely, the embeddings of these textual descriptions are used to train an augmentation model which learns a transformation on the CLIP image embeddings of the training domain and ``places'' them in the new domain (see Figure~\ref{fig:teaser}). We train this augmentation model with two objectives: (1) translating the image embedding from the training domain to the unseen testing domain, while (2) retaining the class-specific information of the original image. Once this transformation is learned, we train a simple linear classifier on the combined augmented and unaugmented image embeddings, resulting in a classifier that outperforms common fine-tuning methods on the extended domain while achieving similar performance on the training domain. 


We introduce \method{}, a method to extend a model to new domains given only a language description of the distribution shift.
Our main contributions are (1) the introduction of the \emph{Domain Extension with Language} problem, (2) a novel language-guided \emph{latent feature augmentation} training procedure, and (3) the extension of our method to address spurious correlation biases in the training data. 

We evaluate \method{} on two domain adaptation benchmarks, DomainNet~\citep{peng2019moment} and CUB-Paintings~\citep{PAN_20}, as well as two benchmarks exhibiting color and contextual bias,  Colored MNIST~\citep{ivm} and Waterbirds~\citep{sagawa2019distributionally}. 
On the domain adaptation benchmarks, we show that we improve out-of-domain performance by 1-3\% while matching in-domain performance of fine-tuned and ensembled models. On the biased benchmarks, we show an almost 2x improvement in out-of-domain performance over fine-tuned models. Across all benchmarks, \method{} achieves the highest accuracy on the entire extended test domain containing both training and unseen test domain samples. Finally, we perform an in-depth analysis of the altered image embeddings, the effect of each loss function, and the effect of different vision and language models to understand our framework better. 

\section{Related Work}
\label{sec:related_work}


\textbf{Domain Adaptation/Generalization.} 
The challenge of out-of-domain generalization is well studied \citep{pmlr-v97-recht19a, gals, kumar2022finetuning, santurkar2021editing,hendrycks2019robustness} with a large body of work in domain adaptation addressing the problem of adapting a model to perform well on a new target domain. A typical domain adaptation approach involves collecting additional unlabeled data from the target domain~\citep{dann, mcd, ivm, lntl, tzeng2015simultaneous}, and aims to train a classifier such that it cannot tell the difference between source and target domain.

In the limited data setting, few-shot domain adaptation \citep{few_shot_da, multi_source_few_shot_da} aims to learn from as little as one example in the target domain.  Work in domain generalization \citep{zs_da, gulrajani2020search,koh2021wilds} does not need target domain data but requires a set of several aligned and labeled source domains, and often shows only limited gains. While we evaluate on certain domain adaptation benchmarks, DA/DG methods primarily focus on maximizing target domain accuracy, while our work is interested in maximizing the accuracy of the \emph{extended domain}. Furthermore, unlike previous works, we assume we have no access to any target data (labeled or unlabeled), only a single source domain, and our domain shift can be verbalized. 

\textbf{Fine-tuning under Distribution Shift. }
The goal of fine-tuning under distribution shift is to tailor pretrained models to a specific task without sacrificing their ability to deal with distribution shifts. \cite{kumar2022finetuning} found that it is better to fit a linear probe on the features and then fine-tune the model's backbone. For robust fine-tuning of CLIP specifically, \cite{robust_clip_ft} proposed ensembling the weights of the fine-tuned image encoder with the zero-shot image encoder. We see our work as complementary to these ideas, targeting semantically defined domain shifts to increase OOD performance, while maintaining high ID performance.

\textbf{Semantic Augmentation with CLIP.}
With the emergence of CLIP, several works~\citep{dalle2, styleclip, gal2021stylegannada} have used language to alter images using a combination of CLIP and a generative model. Broadly, these works translate an image to a CLIP embedding, alter the image embedding with a text embedding of the desired augmentation, and use that embedding to generate an altered image. These CLIP-based works do not attempt to use these data augmentations in the context of dataset bias or domain adaptation. Some prior work has explored augmentations using generative models~\citep{hendricks_tyranny,gan_dom_adaptatino,source_free_da_gan}, but since they generate images at the pixel level, they are often bottle-necked by the quality of the generative process. In contrast, we choose to manipulate embeddings directly that allows us to effectively distill the knowledge in CLIP.  

\textbf{Removing Dataset Bias.}
In computer vision, several works debias data using extra information such as instance annotations~\citep{hendricks2018women, where_to_look, pmlr-v119-rieger20a}, bounding boxes \citep{choi2019sdn}, or image-level bias annotations~\citep{lntl}. 
Some methods~\citep{hendricks_tyranny, rewriting_generative, santurkar2021editing} forego the need for expensive annotations by utilizing generative models, while~\cite{gals} utilize CLIP to translate language descriptions of a task into spatial guidance.
In contrast, we do not limit ourselves to purely spatial bias or use per-image annotations of the bias, only a description of what biases may appear in the training data. 

\label{sec:method}

\section{Latent Augmentation using Domain Descriptions}
We consider the supervised learning problem of generalizing to new unseen domains using only the verbal descriptions of the training domain and the anticipated but unseen new domains.
More formally, we are given a training dataset $\{{\bf x_i}, y_i\}_{i=1}^n$ drawn from the training domain $\dtrain$, the class names $\ctext_y$, a written description $\ttrain$ of the training domain, and a set of written descriptions $\{\tnew^i\}_{i=1}^k$ of $k$ unseen domains $\{\dnew^i\}_{i=1}^k$ that we expect to encounter at test time. 
Our goal is to train a model that performs well on both the original domain $\dtrain$ as well as the unseen domains $\{\dnew^i\}_{i=1}^k$.
We call this the \emph{Domain Extension with Language} problem.

Large vision-language models have demonstrated the ability to generalize to new domains with language but only in the zero-shot setting. In order to utilize available training data, we explore the popular fine-tuning technique of \emph{linear probing}: fitting a linear classifier to the image embeddings of large vision-language models. We chose linear probing over full fine-tuning as it is faster to train and has been shown to result in more robust classifiers~\citep{kumar2022finetuning, Radford2021LearningTV}.

\begin{figure}[t]
    \centering
    \includegraphics[width=0.9\textwidth]{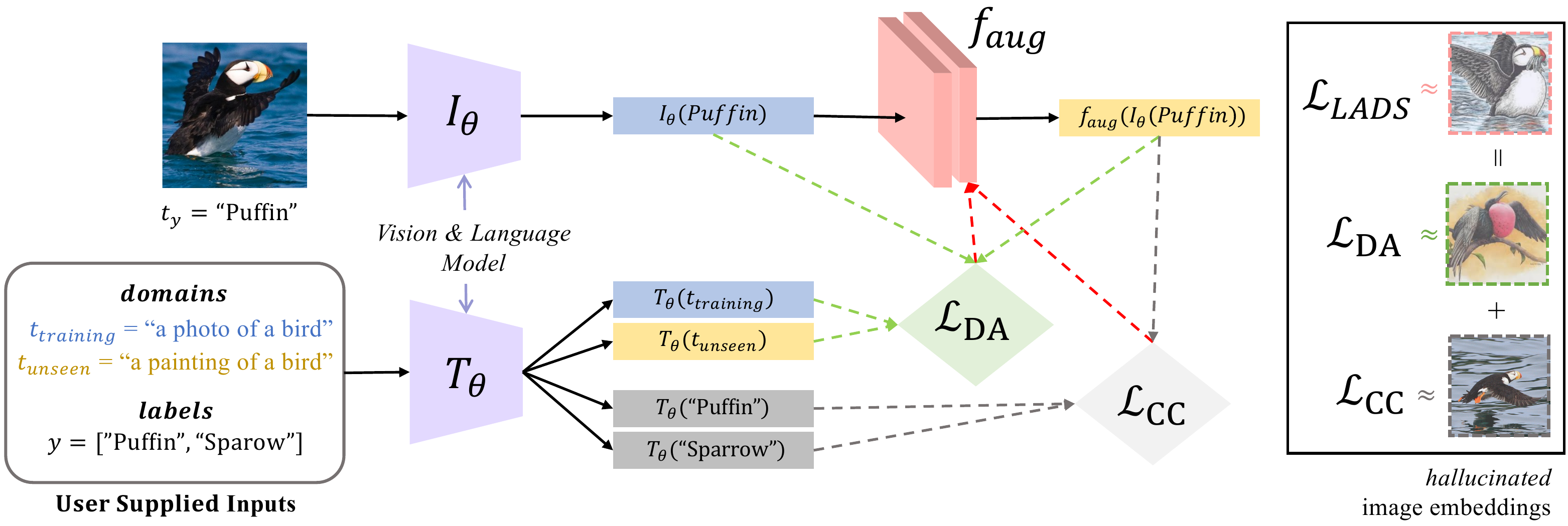}
    \caption{\textbf{\method{}}. Let the task be to classify \emph{Puffin} vs. \emph{Sparrow}. The training data $\dtrain$ contains \emph{photos} of the two classes but we would like to extend our classifier to \emph{paintings} as well: that is, $\dnew$. We aim to do this using the text descriptions of the training and new domain, $\ttrain$ and $\tnew$, respectively. The augmentation network $\faug$ is trained to transform image embeddings from $\dtrain$ to $\dnew$ using a \emph{domain alignment} loss $\mathcal{L}_{\text{DA}}$ and a \emph{class consistency} loss $\mathcal{L}_{\text{CC}}$. When $\mathcal{L}_{\text{DA}}$ is low, the augmented embeddings are in the new domain but may have drifted from their class. When $\mathcal{L}_\text{CC}$ is low, the augmented embeddings will retain class information but may fail to reflect the desired change in domain. $\faug$ aims to augment every image embedding to a space with low domain alignment loss $\it and$ low class consistency loss, resulting in $\faug(I(\bf{x}))$ having an image embedding similar to a painting of a Puffin. Note that the hallucinated image embeddings on the right are a pictorial representation of the effect of each loss function and not actually generated by \method{}.}
    \label{fig:lads}
\end{figure}

While standard linear probing only uses the image embeddings and the numerical labels, \method{} also utilizes the text describing the classes and the descriptions of domain shift to augment the probe's training data to mimic samples from the unseen domain.  
Our two-stage approach first learns a network that transforms the \emph{image embeddings} rather than the pixels themselves, with the goals of (1) augmenting the embedding to be aligned with the unseen domain while (2) retaining the features consistent with its class label. The second stage performs linear probing on the training set containing both the original image embeddings as well as the augmented image embeddings to produce a classifier that is more robust to the specified domains. Note that we do not use any data from $\dnew^k$ in either stage---we only use the class names and domain descriptions. An outline of the first stage of our method (training the augmentation network) is depicted in Figure~\ref{fig:lads}.

We choose CLIP~\citep{Radford2021LearningTV} as our vision-language model in our evaluations. Let $\img{\bf{x}} = \clipimg{\bf{x}} \in \mathcal{I}$ denote the image embedding of input image $\bf{x}$ and $\txt{t} = \cliptext{t} \in \mathcal{T}$ denote the CLIP text embedding of some text $t$. Furthermore, let $\ttrain \circ \ctext_y$ denote the composition of the domain description and the class name. For example, if $\ttrain =$ \emph{``a photo of a''}, $t^1_\text{unseen} =$ \emph{``a painting of a''} and $\ctext_y$ could be \emph{``Puffin''}. The composition $\ttrain \circ \ctext_y$ is \emph{``a photo of a Puffin''}.

\paragraph{Stage 1: Training the augmentation network.} The first stage of \method{} is to learn an augmentation network $\faug^k:\mathcal{I} \to \mathcal{I} $ that transforms image embeddings from $\dtrain$ to $\dnew^k$ using the corresponding language descriptions $\ttrain$ and $\tnew^k$. 
As mentioned previously, a valuable augmentation is one which places the transformed embedding in unseen domain $\dnew^k$ while retaining features relevant to the class label. 
To achieve this, we train $\faug^k$ using a combination of two losses: \emph{Domain Alignment} and \emph{Class Consistency}. In the setting of adapting to multiple new domains at once, we train a unique $\faug^k$ network for each domain as described above. 

\textit{Domain Alignment.} The domain alignment loss encourages the augmented image embeddings $\faug^k(\img{\bf{x}})$ to look like image embeddings from the new domain $\dnew^k$. This loss is guided by the text embeddings of the domain descriptions $\tnew^k$ and $\ttrain$. 

While CLIP is trained such that the space of image embeddings $\mathcal{I}$ has some correspondence with the space of text embeddings $\mathcal{T}$, it is not obvious what a mapping between $\mathcal{I}$ and $\mathcal{T}$ should look like. Thus, inspired by prior work ~\citep{styleclip, gal2021stylegannada}, we assume the existence of a ``global direction'' that corresponds to a shift from $\dtrain$ to $\dnew^k$ that is shared across both the image embedding space and text embeddings space. 

This ``global direction'' is defined as the normalized difference of the embeddings from the target domain and the embeddings from the source domain. Formally, the domain alignment loss of $\faug ^k$ for training point $({\bf x_i}, y_i)$ is
\begin{align}
\label{eq:da}
    \mathcal{L}_{\text{DA}}(\faug^k) = \sum \limits_{i=1}^n 1 - \left(\frac{\faug^k(\img{{\bf{x_i}}}) - \img{{\bf{x_i}}}}
    {\| \faug^k(\img{{\bf{x_i}}}) - \img{\bf{x_i}} \|} \cdot \frac{\txt{\tnew^k, y_i} - \txt{\ttrain, y_i}}{\| \txt{\tnew^k, y_i} - \txt{\ttrain, y_i} \|}
    \right).
\end{align}

\textit{Class Consistency.} The domain alignment loss in~\Eqref{eq:da} encourages the augmented embeddings to only differ in the direction of change in the domain. If there were one global shared direction corresponding to the domain shift, optimizing $\mathcal{L}_\text{DA}$ would be sufficient. 
However, in practice, we find that optimizing $\mathcal{L}_\text{DA}$ alone removes some class relevant information and results in little diversity among the augmented embeddings of different images (see Section~\ref{sec:loss_ablation} and Figure~\ref{fig:cub_nn_da}). 
Thus we add a class consistency loss which preserves class information in the augmented embeddings. 
We measure class information in the image embeddings by our ability to classify the images accurately via CLIP zero-shot with the class names. Formally, 

\begin{align}
\label{eq:cc}
    \mathcal{L}_{\text{CC}}(\faug^k) =
    \sum \limits_{i=1}^n \text{Cross-entropy}\big(\text{Softmax}[\faug^k(\img{{\bf{x_i}}}) \cdot \txt{y_i}], y_i \big)
\end{align}

Note that this is the same objective as the standard CLIP loss. We use CLIP zero-shot rather than trying to fine-tune CLIP because that could lead to overfitting where we classify the augmented training image embeddings correctly even when they do not contain class relevant information.

Our final objective $\mathcal{L}_\text{LADS}(\faug)$ to train the augmentation network as the first step in LADS is a linear combination of the the domain alignment loss and class consistency loss:
\begin{align}
\mathcal{L}_\text{LADS}(\faug^k) = \alpha \mathcal{L}_\text{DA}(\faug^k) + (1-\alpha) \mathcal{L}_\text{CC}(\faug^k),
\end{align}where $\alpha$ is a hyperparameter dictating the trade-off between domain alignment and class consistent.

\paragraph{Stage 2: Fine-tuning.} After the augmentation network $\faug ^k$ is trained, we train a linear probe on the original image embeddings $\img{{\bf x_i}}$ along with the augmented embeddings $\faug^k(\img{{\bf x_i}})$. Inference is straightforward: apply the linear probe on the CLIP image embeddings of the test images.


\subsection{Addressing dataset bias}
In addition to dealing with extended domains, \method{} can also be used in the dataset bias setting where there are spurious correlations in the dataset. For example, in Waterbirds~\citep{sagawa2019distributionally}, we want to classify Landbirds vs. Waterbirds, where the spurious correlation is the background (Landbirds appear on forest backgrounds and Waterbirds appear on water backgrounds in training). To prevent a classifier from using this correlation to make predictions, we can use \method{} to generate augmentations that represent ``Landbird on water'' and ``Waterbird on land''. 

We do this by using CLIP to label the backgrounds of each image and then decide what $\ttrain$ and $t_\text{unseen}$ is per example. Given the domain information $t_{land}$ = \emph{``a \{\} in the forest''} and $t_{water}$ = \emph{``a \{\} on the water''}, we can use zero-shot CLIP to determine if a given image is on land or water. If the image is predicted to be on land, when training $\faug$, $\mathcal{L}_{\text{DA}}$ for that particular example will use $\ttrain = t_{land}, t_\text{unseen} = t_{water}$ and vice versa. The class consistency loss and the other parts of the pipeline remain unchanged. Because we are using the vision and language model to label the domains, we do not need per-image labels of the bias, only a hypothesis of what the bias may be. 


\section{Experiments}
\label{sec:experiments}
In this section we discuss our main experiments and results. We defer dataset details, the remainder of the experiments and their discussion to the Appendix (\ref{supp:datasets}, \ref{supp:visual}, \ref{supp:mnist-svhn-appendix}).

\subsection{Implementation Details}
\label{sec:implementation}
In line with ~\cite{Radford2021LearningTV}, we normalize all text and image embeddings when performing zero-shot inference or training with CLIP embeddings. The augmentation network $\faug$ used in \method{} is a 2-layer MLP with input and output dimensions of 768 and a hidden dimension of 384. Within \method{} and all the CLIP-related baselines, we use the OpenAI CLIP model with a ViT-L backbone and resize all images to 224x224. We train on 10 GeForce RTX 2080 Ti GPUs.

For each baseline, we do a hyperparameter sweep across learning rate and weight decay and choose the parameters with the highest class-balanced validation accuracy. For \method{} we also do a sweep across the parameters of the augmentation network, namely learning rate, weight decay, and $\alpha$, and select a checkpoint based on the validation loss. In general, we set $\alpha=0.5, lr=0.001, wd=0.05$. Our hyperparameter search spaces and final choice of hyperparameters are listed in Table~\ref{tab:dataset_stats}. 

In our results we report test accuracy on $\dtrain$, $\dnew$, and the extended domain which averages the two. We run each method over 5 different random seeds and report the mean and standard deviation.

\subsection{Datasets}
\label{sec:datasets}

\textbf{CUB-Paintings (one new domain)} is composed of 2 datasets, CUB-200~\citep{Wah11thecaltech-ucsd}, a fine-grained bird classification benchmark containing 200 different bird species and CUB-200-Paintings~\citep{PAN_20}, which contains the same classes as CUB-200 but instead of real images they are paintings 
collected from the web and filtered manually. We use the domain descriptions $\ttrain=\text{``a photo of a \{\} bird"}, \tnew^1=\text{``a painting of a \{\} bird"}$.

\textbf{DomainNet (multiple new domains)} is a specific split~\citep{domainnet_split} of the original DomainNet~\citep{peng2019moment} dataset which contains the 40 most common classes from 4 domains: `sketch', `real', `clipart', and `painting'. Like prior work~\citep{kumar2022finetuning, domainnet_split}, we train on sketches and evaluate on the three other domains. We use the domain descriptions $\ttrain=\text{``a sketch of a "}, \tnew^1=\text{``clipart of a "}, \tnew^2=\text{``a painting of a "}, \tnew^3=\text{``a realistic photo of a "}$.

\textbf{Colored MNIST (color bias)}~\citep{ivm} was made by taking the original MNIST Digits~\citep{deng2012mnist}, and coloring them red or blue. In the training and validation sets, even numbers are red and odd numbers are blue, while in the test set digits are colored randomly. The task is to classify the digits $0,1,..,9$. 
We use the domain descriptions $\text{``a photo of a red number "}, \text{``a photo of a blue number "}$. 

\textbf{Waterbirds (contextual bias) }~\citep{sagawa2019distributionally} is a synthetically created dataset which creates contextual bias by taking species of landbirds and waterbirds from the CUB-200~\cite{Wah11thecaltech-ucsd} dataset and pasting them on forest and water backgrounds from the Places~\citep{zhou2017places} dataset. For the training and validation sets, all landbirds appear on forest backgrounds and waterbirds appear on water backgrounds while the test set has an even representation of backgrounds and bird types. We use the domain descriptions $\text{``a photo of a \{\} in the forest"}, \text{``a photo of a \{\} on the water"}$. 

\subsection{Baselines}
\label{sec:baselines}
\emph{Generic and Adaptive zero-shot CLIP} are the zero-shot baselines proposed by ~\cite{Radford2021LearningTV}: (CLIP ZS (G)) uses the class name alone as the text prompt, while adaptive zero-shot CLIP (CLIP ZS (A)) caters the text prompts to the specific domains (e.g. ``a painting of an airplane.''). To do well on the extended domain, we average the text embeddings of each class across all possible domains. 

\emph{CLIP LP} fits a linear classifier on top of the CLIP image embeddings.

\emph{CLIP LP (ZS init)} initializes the linear classifier with the text embeddings.

\emph{WiSE
-LP}~\citep{robust_clip_ft} is an ensembling technique which fine-tunes a CLIP model and does a weighted average of the fine-tuned model's weights with the original. Due to the size of the vision and language models we are using, we did not fine-tune the entire backbone and instead ensembled the classifier with the linear classifier probe as explained by~\cite{robust_clip_ft}.

\emph{{VQGAN + CLIP}}~\citep{vqgan_clip} is a method that uses a VQGAN~\citep{esser2021taming} trained with CLIP to augment images in pixel space. Using a text prompt and an image, we perform ``style transfer'' to the new domain in order to augment the training data. We then train a linear probe on the augmented and non-augmented CLIP embeddings. Due to the amount of time and compute required to generate images, we only ran this baseline for DomainNet and augmented approximately 15\% of the training dataset. Examples of the augmented images are provided in Table~\ref{fig:vqgan_vis}. 


\subsection{Results}
\label{sec:results}

Table~\ref{tab:cub} shows in-domain (ID) and out-of-domain (OOD) accuracy on CUB-Paintings and DomainNet. The ``Extended'' column is the average accuracy of the two, corresponding to the full extended domain.
For CUB-Paintings and DomainNet, \method{} is able to match or improve the ID accuracy of the fine-tuning baselines while improving over their OOD accuracy. Although CLIP zero-shot achieves higher OOD accuracy on DomainNet, \method{} achieves the highest result when evaluated on the full extended domain. We also improve over the VQGAN+CLIP baseline on DomainNet. 

\begin{figure}[h]
     \centering
     \begin{subfigure}[b]{\textwidth}
         \centering
         \includegraphics[width=0.9\textwidth]{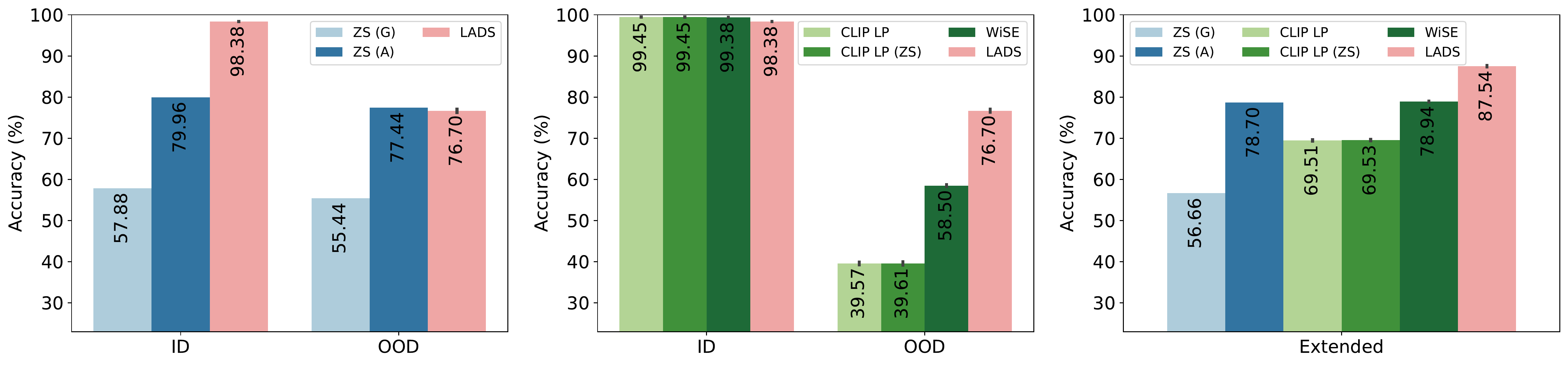}
         \caption{\textbf{Colored MNIST.}}
         \label{fig:colored_mnist}
     \end{subfigure}
     \hspace{1cm}
     \begin{subfigure}[b]{\textwidth}
         \centering
         \includegraphics[width=0.9\textwidth]{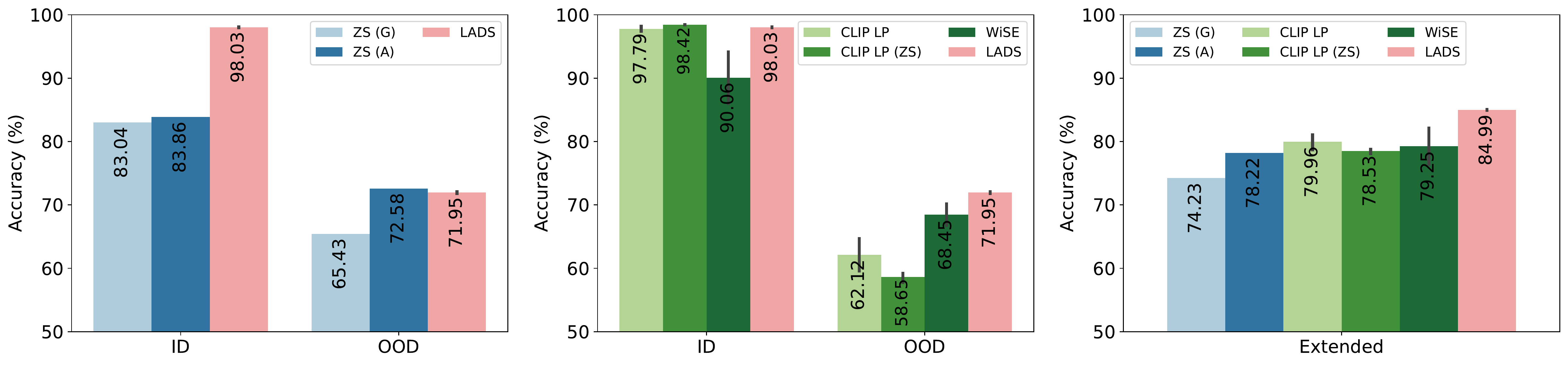}
         \caption{\textbf{Waterbirds.}}
         \label{fig:waterbirds}
     \end{subfigure}
        \caption{\textbf{Result on dataset bias benchmarks.} Left and center plots show the training domain and unseen domain performance of the zeroshot and fine-tuned baselines respectively. For both Colored MNIST (a) and Waterbirds (b), \method{} is able to roughly match the unseen domain accuracy of zero-shot methods and the seen domain accuracy of fine-tuned methods, resulting in improved performance on the extended domain (right).}
        \label{fig:bias_datasets}
\end{figure}

For Colored MNIST (Figure~\ref{fig:colored_mnist}) and Waterbirds (Figure~\ref{fig:waterbirds}), \method{} is able to roughly match ID accuracy of the fine-tuned CLIP and OOD accuracy of CLIP zero-shot, resulting in approximately a $10\%$ improvement on the extended domain. We explore different weighted averages of ID and ODD accuarcy to compute the extended domain accuracy in Section~\ref{supp:extended_tradeoffs} of the Appendix.

\begin{table}[h]
\small
    \centering
    \begin{tabular}{lllll}
    \toprule
    Dataset & Method     & ID & OOD & Extended \\
    \midrule
    CUB-Paintings & CLIP ZS (G) & 60.34\% & 52.84\% & 56.59\% \\
    CUB-Paintings & CLIP ZS (A) & 61.93\% & 54.38\% & 58.16\% \\
    \midrule
    CUB-Paintings & CLIP LP     & \textbf{85.91$\pm$0.08\%} & 64.33$\pm$0.29\%  & 75.12$\pm$0.18\% \\
    CUB-Paintings & CLIP LP (ZS init)     & \textbf{86.08$\pm$0.11\%} & 65.05$\pm$0.05\%  & 75.57$\pm$0.06\% \\
    CUB-Paintings & WiSE-LP & 81.74$\pm$0.34\% & 64.80$\pm$0.10\% & 73.27$\pm$0.22\% \\
    CUB-Paintings & \method{}     & \textbf{86.14$\pm$0.29\%} & \textbf{66.18$\pm$ 0.25\%} & \textbf{76.16$\pm$0.23\%} \\
    \midrule[2pt]
    DomainNet & CLIP ZS (G) & 93.49\% & 95.94\% & 94.72\% \\
    DomainNet & CLIP ZS (A) & 93.24\% & \textbf{96.01}\% & 94.62\% \\
     \midrule
    DomainNet & CLIP LP & 95.03$\pm$0.07\% & 93.75$\pm$0.02\% & 94.39$\pm$0.04\%\\ 
    DomainNet & CLIP LP (ZS init) & \textbf{95.21$\pm$0.21}\% & 93.95$\pm$0.03\% & 94.58$\pm$0.11\% \\
    DomainNet & WiSE-LP     & 95.19$\pm$ 0.34\% & 93.68$\pm$ 0.12\%& 94.44$\pm$0.11\% \\
    DomainNet & VQGAN+CLIP & \textbf{95.54$\pm$ 0.09}\% & 93.83$\pm$ 0.10\% & 94.67$\pm$ 0.09\%\\
    DomainNet & \method{}     & \textbf{95.33 $\pm$ 0.33}\% & 95.21 $\pm$ 0.09\% & \textbf{95.27$\pm$ 0.14}\%\\
    \bottomrule
  \end{tabular}
  \vspace{2mm}
  \caption{In-domain (ID), out-of-domain (OOD) and extended domain accuracy on \textbf{CUB-Paintings} and \textbf{DomainNet}. For DomainNet, we include the pixel augmentation baseline VQGAN+CLIP and OOD accuracy is the average of the 3 unseen domains. \method{} is able to beat all methods on the extended domain for both datasets. Note that for tasks where CLIP zero-shot does not perform well, \method{} is able to significantly outperform zero-shot on the unseen domain.} 
  \label{tab:cub}
\end{table}


\subsection{Analysis of Augmentation Quality}
\label{sec:aug_analysis}
In this section, we explore the quality of the embeddings generated by our augmentation network. We perform our analysis for DomainNet below, defering the remaining results to Appendix~\ref{supp:nn_visual},~\ref{supp:aug_quality}.

Since there is no publicly available model to convert CLIP embeddings to images, we use the nearest neighbors of the augmented embeddings from the extended test domain to confirm that our augmentations match our expectations. We take a random subset of 1,000 samples from the image embeddings used to train the linear probe: for CLIP LP, this is simply $\{{\img{\bf x_i}}, y_i\}_{i=1}^n$, for VQGAN+CLIP it is of a mix of $\{{\img{\bf x_i}}, y_i\}_{i=1}^n$ and GAN generated images, and for \method{} it is $\{{\img{\bf x_i}}, y_i\}_{i=1}^n$ and the augmented embeddings $\bigcup_{j=1}^k\{{\faug^j(\img{\bf x_i}}), y_i\}_{i=1}^n$ for each unseen domain $j$. We obtain the nearest neighbors in the extended test set (containing images from the training and unseen domain) with respect to cosine similarity of the image embeddings. 
\\
In line with our domain alignment and class consistency loss functions, we define metrics for (1) correctly altering the domain of the image embedding, while (2) retaining the class information. We define the percentage of the nearest neighbors that belong in the desired domain as the \textit{domain alignment score}, and the percentage that belong to the original class as the \textit{class consistency score}.



\begin{table}
\small
    \centering
  \begin{tabular}{llll}
    \toprule
         & CLIP LP & VQGAN+CLIP & \method{}  \\
    \midrule
    Domain Alignment score    & 81.30$\pm$1.35\% & 69.26$\pm$2.31\% & 85.44$\pm$0.61\% \\
    Class Consistency score& 91.42$\pm$0.47\% & 77.14$\pm$1.22\% & 87.26$\pm$0.74\%\\
    \bottomrule
  \end{tabular}
  \vspace{2mm}
        \caption{\textbf{Augmentation Quality for DomainNet.} The domain alignment and class consistency scores over 1000 randomly sampled training embeddings and their nearest neighbor in the test set. \method{} achieves higher domain alignment and class consistency scores than VQGAN+CLIP, and is able to obtain a somewhat similar class consistency score to the unaugmented image embeddings. }
        \label{fig:nn_metrics}
\end{table}

The CLIP LP scores can be viewed as an approximate upperbound for those of \method{} since they reflect the nearest neighbors of only the original sketch embeddings in the extended domain.
As shown in Table~\ref{fig:nn_metrics}, \method{} is able to beat the domain alignment score and closely fall behind the class consistency score of the linear probe, implying that the augmentations are of similar quality to the original image embeddings. Furthermore, \method{} has better domain alignment and class consistency than VQGAN+CLIP, indicating that the long and laborious pixel-level augmentation may be producing lower quality training samples than our simple embedding augmentation. 

\begin{figure}[ht]
    \centering
    \includegraphics[width=\textwidth]{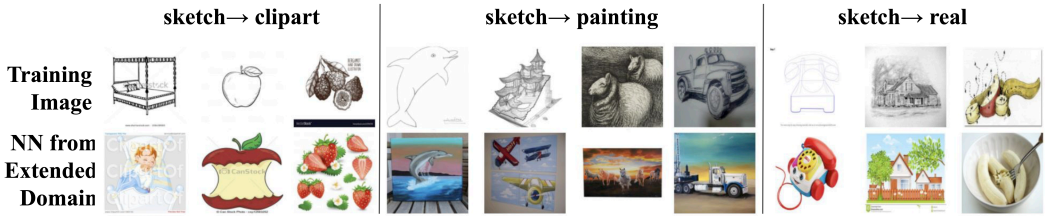}
    \caption{\textbf{Nearest Neighbors for \method{} on DomainNet Sketch $\to$ Clipart, Painting, Real.} The top row shows training images with the label on top being the intended domain augmentation for that embedding. The bottom row shows the nearest neighbor of the augmentation in the extended domain. Not only does \method{} produce augmented embeddings within the correct domain, embeddings often match the specific class and stylistic elements of each original image. }
    \label{fig:nn_vis}
\end{figure}

For qualitative analysis of \method{}, we visualize a random sample of 10 nearest neighbors from DomainNet in Figure~\ref{fig:nn_vis} (the sketch embeddings are non-augmented, all others are augmented). The nearest neighbors of augmented embeddings closely resemble embeddings of similar images in the desired unseen domain. Even if the nearest neighbor is of a different class, it maintains some visual similarity to the original image. 

\subsection{Ablations}
In this section, we ablate the class consistency and domain alignment loss described in Section~\ref{sec:method}. We defer the remainder of the ablations, including ablations of the domain descriptions and the CLIP model, to Appendix~\ref{supp: nn_ablate_loss},~\ref{supp:robustness wrt prompts-appendix},~\ref{supp:clip_ablation}.

\label{sec:loss_ablation}

In order to measure the impact of the domain alignment and class consistency loss functions, we ablate each one and report the accuracy, domain alignment score, and class consistency score. We also experiment with a domain-specific class consistency loss, which replaces $T(y_i)$ with $T(\tnew^k \circ y_i)$ in order to enforce the class and domain all in one loss. We display our results on the Waterbirds dataset below, with experiments on the other datasets in Appendix~\ref{supp: nn_ablate_loss}. 

\begin{figure}[ht]
    \begin{subfigure}{0.49\textwidth}
         \includegraphics[width=\textwidth]{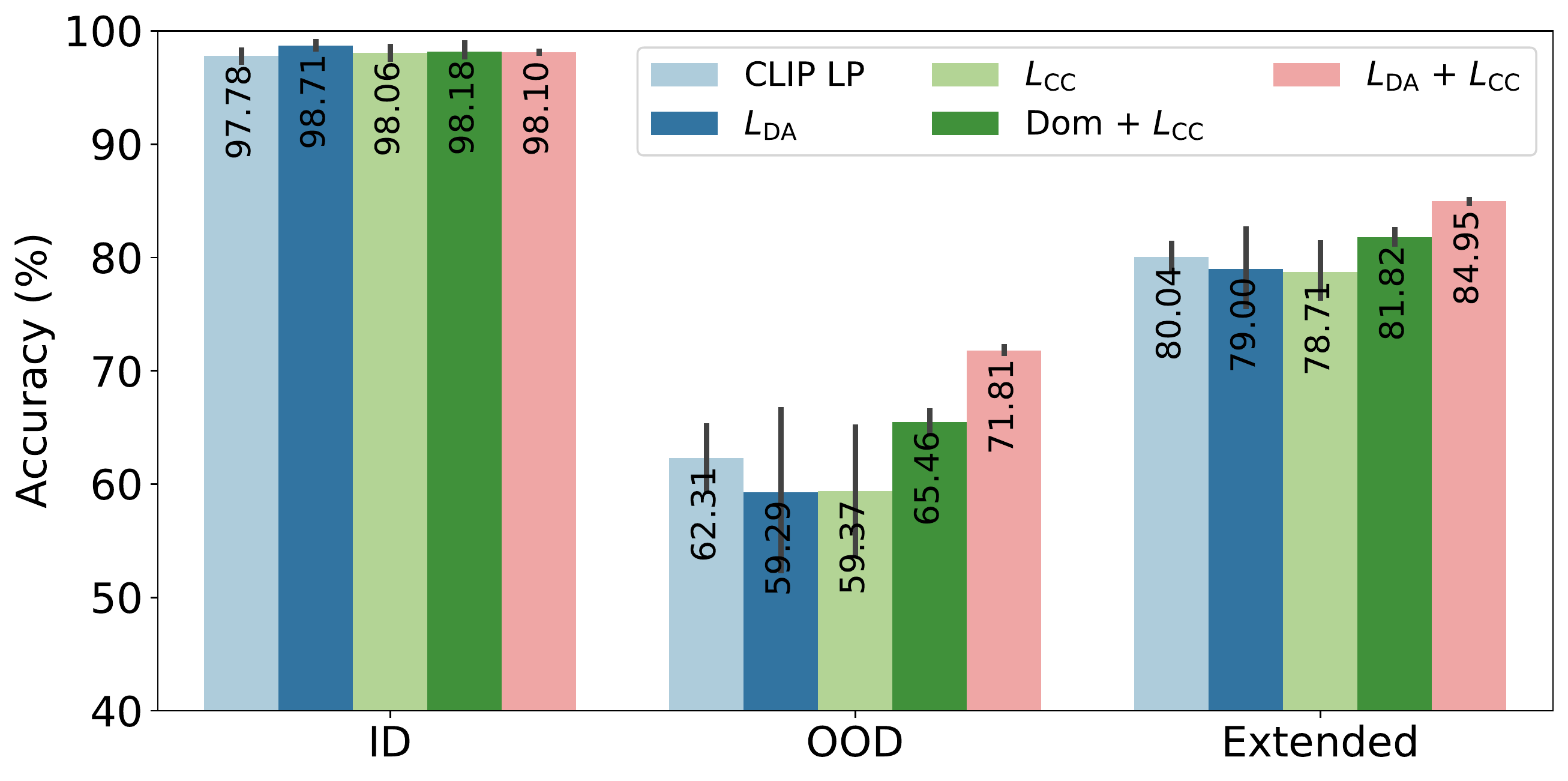}
         \label{fig:loss_ablation_acc}
    \end{subfigure}
    \hspace{1cm}
    \begin{subfigure}{0.30\textwidth}
         \includegraphics[width=\textwidth]{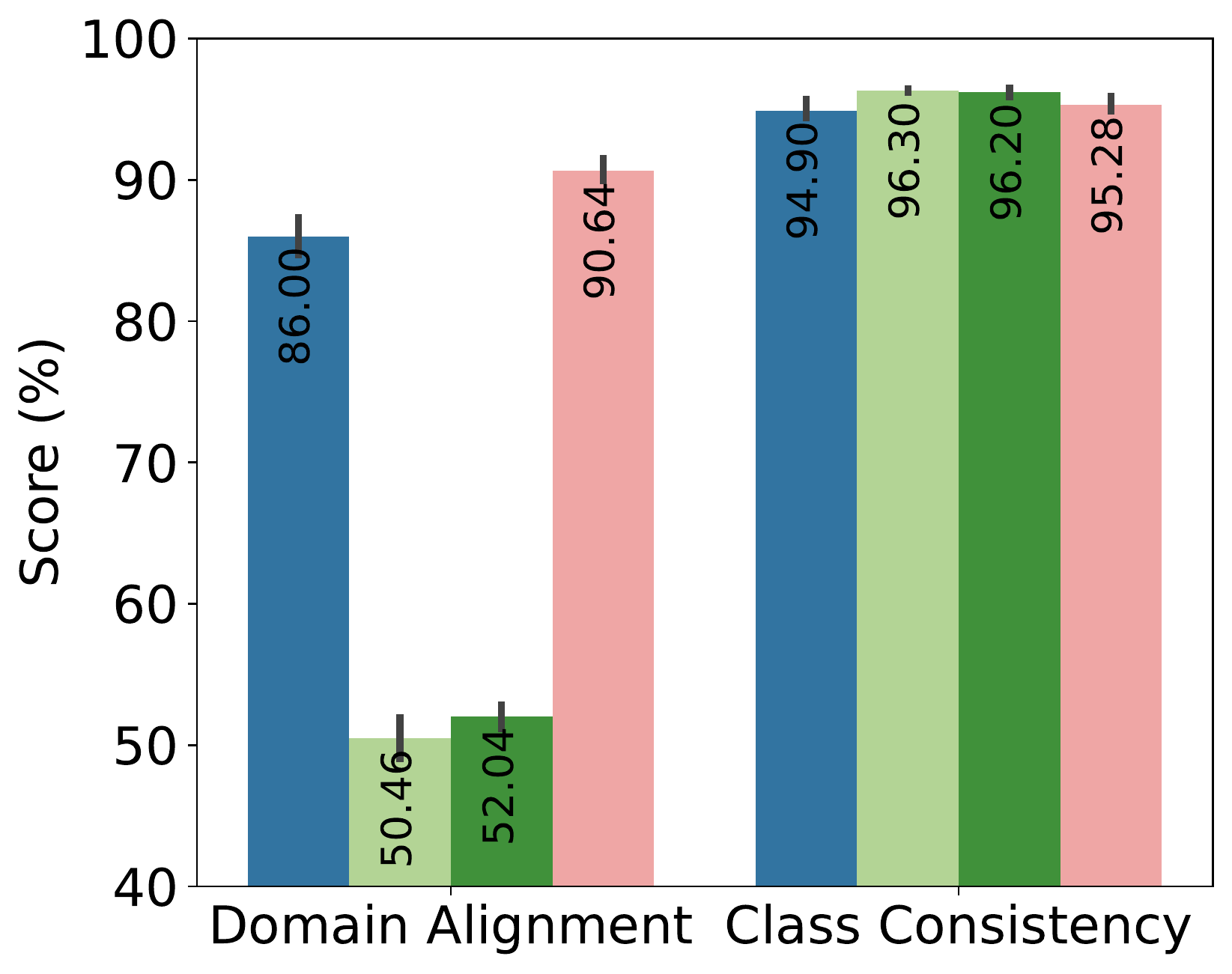}
         \label{fig:loss_ablation_score}
     \end{subfigure}
     \captionsetup{aboveskip=0pt}
     \centering
     \caption{\textbf{Effect of the Loss Functions.} We report the results of training with just the domain alignment loss, the class consistency loss, a domain-specific class consistency loss, and the domain alignment $+$ class consistency loss, on Waterbirds. The DA loss results in high DA score but low accuracy. The CC loss results in low DA score and does not improve the OOD accuracy; the domain-specific CC variant brings negligible gains. Our final design ($\mathcal{L}_{\text{DA}}$+$\mathcal{L}_{\text{CC}}$) works the best.}
    \label{fig:loss_ablation}
\end{figure}

As shown in Figure~\ref{fig:loss_ablation}, the domain alignment loss alone results in a high domain alignment score, but low accuracy due to losing some class specific information.
Meanwhile, the class consistency loss alone achieves the highest class consistency score because it retains the relevant class information, but it fails to improve the OOD accuracy since the augmented embeddings are not within the new domain. Even in the case of domain specific $\mathcal{L}_{\text{CC}}$ when the extended domain is incorporated into the class consistency loss, the scores only slightly improve. It is only when we combine both our losses that we are able to retain class information while transforming the image embeddings to the desired domain, leading to improved out-of-domain accuracy. Nearest neighbor visualizations of the different losses are given in Appendix~\ref{supp: nn_ablate_loss}.

\section{Limitations and Future Work}
\label{sec:limitations}
 Since one must input a natural language description of the distribution shift, \method{} may not apply to ``natural'' distribution shifts where the change cannot be verbalized ~\citep{koh2021wilds}. Furthermore, as our approach is reliant on the richness of concepts learned by a pretrained vision-language model, it is also limited to domains that can be accurately represented with textual descriptions, and are well covered in the data the pretrained models were trained on. As a general rule of thumb, if CLIP zero-shot has very poor performance when it comes to classifying the domains and/or classes, \method{} should not be used (see Section~\ref{supp:mnist-svhn-appendix} of the Appendix). 
%


We have presented \method{}, a fine-tuning method for addressing the task of Domain Extension with Language. We view \method{} as a jumping-off point for further exploration regarding how we can use the zero-shot capabilities of large multimodal models to improve accuracy on a desired domain given only language description as input.
We hope that future work is able to perform reliable embedding augmentations independently of the ability of CLIP to correctly classify the domains and classes at hand. Furthermore, we hope future work is able to analyze more complicated domain shifts such as the ones seen in WILDS~\cite{koh2021wilds} or Imagenet-A~\citep{hendrycks2021nae}.

\textbf{Acknowledgements.}
This work was supported in part by the NSF CISE Expeditions Award (CCF-1730628), DARPA’s SemaFor, PTG and/or LwLL programs, and BAIR’s industrial alliance programs.

\medskip

{
\small
\bibliography{references}
\bibliographystyle{iclr2023_conference}
}


\newpage
\appendix

\noindent\makebox[\linewidth]{\rule{\linewidth}{3.5pt}}
\begin{center}
	\bf{\Large Supplementary Material for \\
``Using Language to Extend to Unseen Domains''}
\end{center}
\noindent\makebox[\linewidth]{\rule{\linewidth}{1pt}}


Here we include experimentation details such as hyperparameters and their search spaces (Section~\ref{supp:hyperparams}), dataset statistics (Section~\ref{supp:datasets}), as well as additional analysis and visualizations (Section~\ref{supp:extended_tradeoffs},~\ref{supp:visual},~\ref{supp:mnist-svhn-appendix}).

\section{Hyperparameters}
\label{supp:hyperparams}

We provide the hyperparameters used in Section~\ref{sec:experiments} of the main paper in Table~\ref{tab:hps}, as well as include additional training details. 

For the linear probing methods, we sweep over learning rates [0.1, 0.05, 0.01, 0.005, 0.001] and weight decay [0.5, 0.05, 0.005, 0.0005]. We also sweep over the same search space when deciding the learning rate and weight decay for the augmentation network (Aug). We train logistic regression on embeddings for 400 epochs. 

\begin{table}[h]
\small
    \centering
    \begin{tabular}{lllllll}
    \toprule
    Dataset & Method & LR & WD & Aug LR & Aug WD & $\alpha$\\
    \midrule
    CUB-Paintings & \method{} & 0.001 & 0.05 & 0.001 & 0.05 & 0.5\\
    CUB-Paintings & CLIP LP & 0.001 & 0.05 & - & - & - \\
    \midrule
    DomainNet & \method{} & 0.0001 & 0.05 & 0.0001 & 0.05 & 0.5 \\
    DomainNet & CLIP LP & 0.0001 & 0.05 & - & - & - \\
    \midrule
    Colored MNIST & \method{} & 0.001 & 0.05 & 0.005 & 0.05 & 1 \\
    Colored MNIST & CLIP LP & 0.005 & 0.05 & - & - & - \\
    \midrule
    Waterbirds & \method{} & 0.001 & 0.05 & 0.005 & 0.05 & 0.75\\
    Waterbirds & CLIP LP & 0.001 & 0.05 & - & - & - \\
    \bottomrule
  \end{tabular}
  \caption{\textbf{Experiment Hyperparameters.} ``Aug'' LR and WD refers to the learning rate and weight decay of the augmentation network used in \method{}. Note that we use $\alpha = 1$ for Colored MNIST since CLIP Zero-Shot does poorly on classifying MNIST.}
  \label{tab:hps}
\end{table}

\section{Dataset Statistics}
\label{supp:datasets}

We provide the training, validation and test (ID and OOD) splits for each dataset in Table~\ref{tab:dataset_stats}.

\begin{table}[h]
\small
    \centering
  \begin{tabular}{lrrrrrrr}
    \toprule
     & \multicolumn{2}{c}{Training} & \multicolumn{2}{c}{Validation}& \multicolumn{2}{c}{Testing} \\
    \cmidrule(r){2-3} \cmidrule(r){4-5} \cmidrule(r){6-7} 
    Dataset     & ID & OOD & ID & OOD & ID & OOD \\
    \midrule
    CUB-Paintings & 5,994 & 0 & 2,897 &0 & 2,897 & 3,047 \\
    DomainNet & 5,537 &0 & 1,200 &0 &   1,200 & 11,468 \\
    Colored MNIST & 30,000 &0 & 5,000&0  & 5,000 & 5,000\\
    Waterbirds &4,795 &0 &600 &0 &2,897 &2,897 \\
    \bottomrule
  \end{tabular}
  \caption{\textbf{Dataset Statistics.} Counts of ID and OOD samples in each split of the data.}
  \label{tab:dataset_stats}
\end{table}

\section{Extended Domain Accuracy vs Amount of ID Test Data}
\label{supp:extended_tradeoffs}
\begin{figure}[h]
    \centering
    \includegraphics[width=\textwidth]{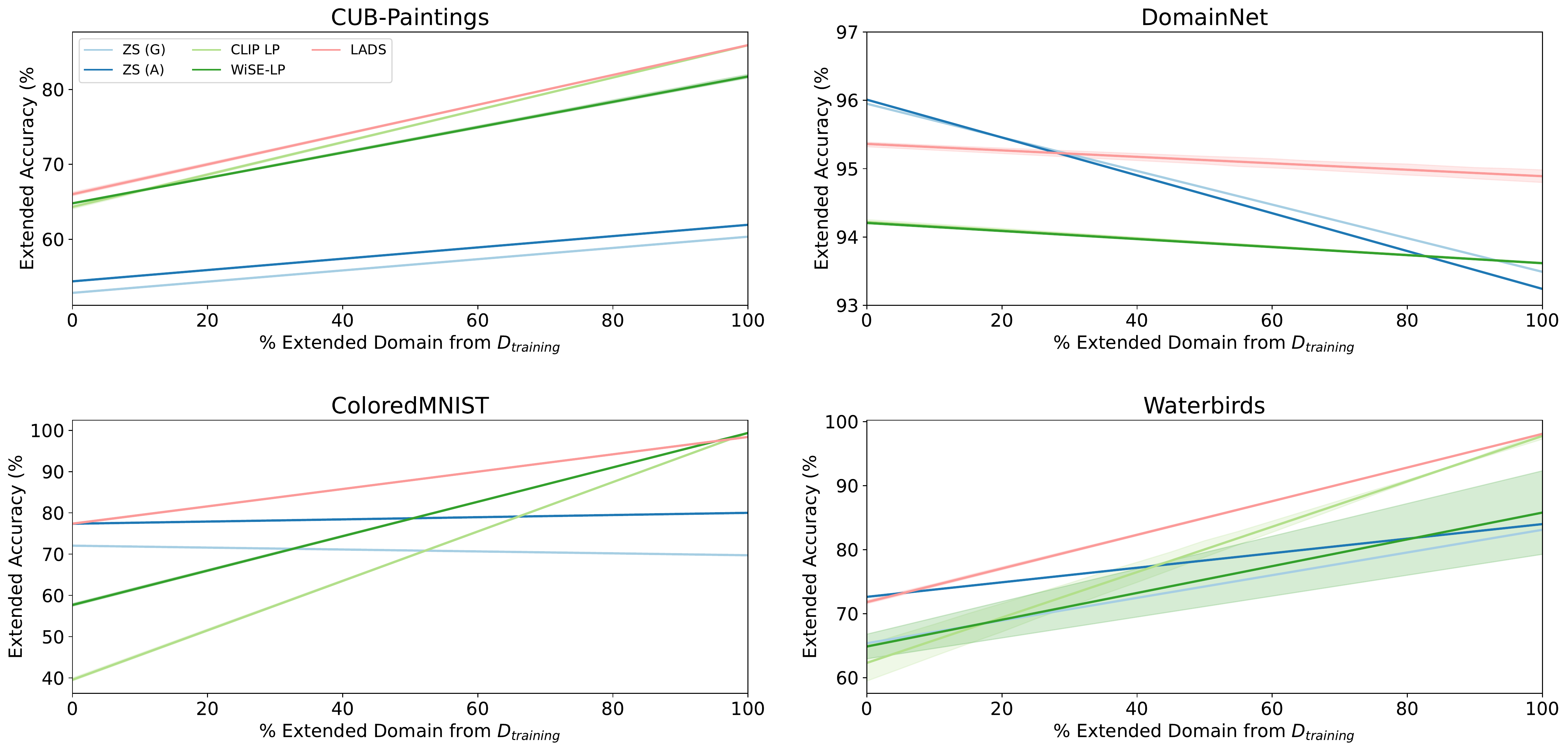}
    \caption{\textbf{Extended Domain Accuracy vs Amount of ID Test Data.} For each dataset, we compute the extended accuracy as a weighted average of the proportion of testing data from $\dtrain$ and $\dnew$. As the proportion of the extended domain belonging to $\dtrain$ increases, the extended domain accuracy of fine-tuning methods and \method{} is more favorable than zero-shot.}
    \label{fig:weighted_extended_domain} 
\end{figure}

While we report extended domain accuracy as the overage of ID and OOD accuracy, we plot the extended domain accuracy as a function of the proportion of data in the extended domain that belongs to $\dtrain$ in Figure~\ref{fig:weighted_extended_domain}. For example, if we assume 70\% of our extended domain is from $\dtrain$, our extended domain accuracy would be $0.7\times$ ID Acc $+$ $0.3\times$ OOD Acc. We can see that as the proportion of the extended domain belonging to $\dtrain$ increases, the extended domain accuracy of fine-tuning methods and \method{} is more favorable than zero-shot. 

\section{Additional Analysis and Visualizations} 
\label{supp:visual}

\subsection{Examples of the augmented images produced by VQGAN+CLIP}
\label{supp:nn_visual}

\begin{figure}[h]
    \centering
    \includegraphics[width=\textwidth]{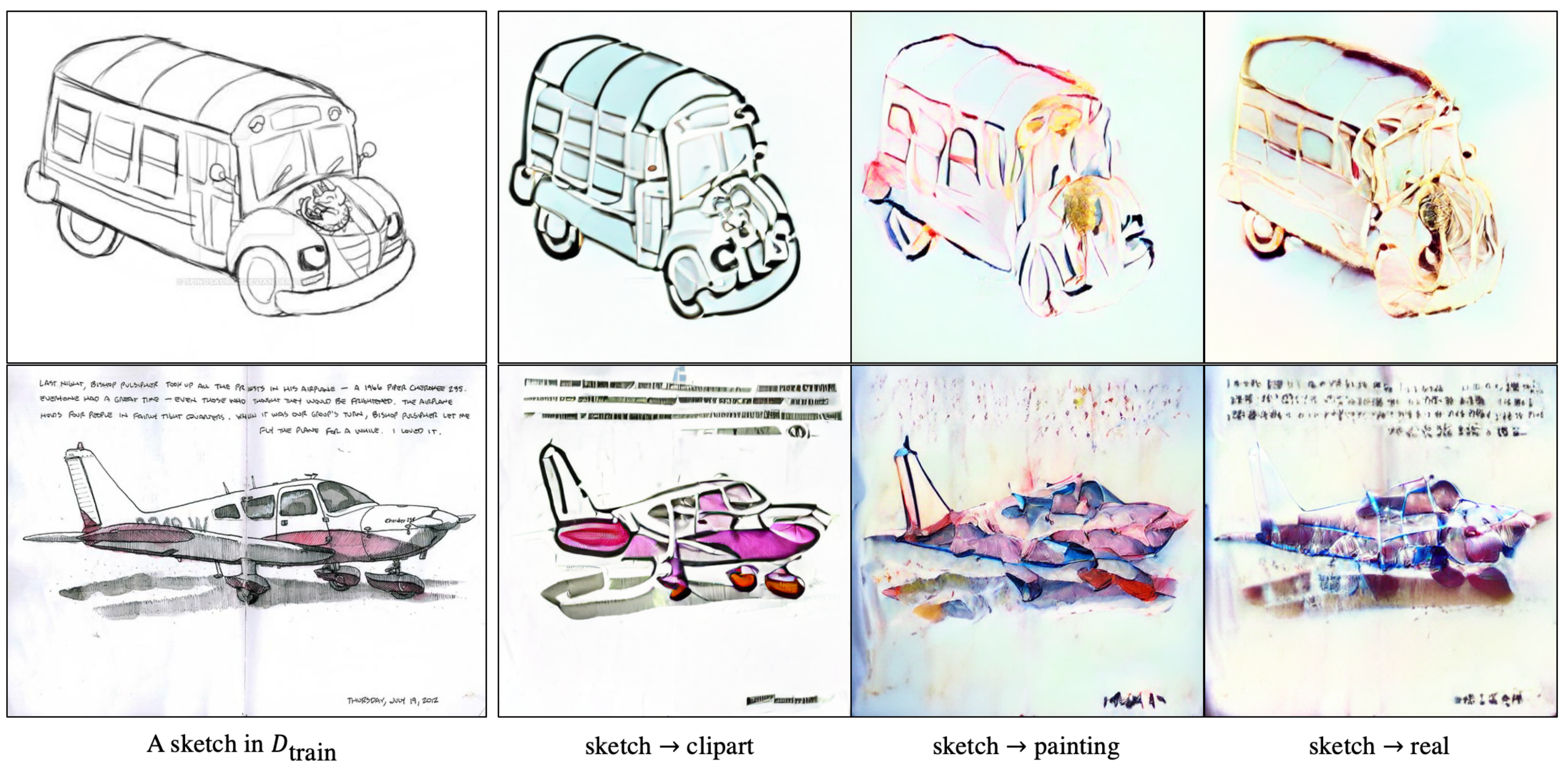}
    \caption{\textbf{VQGAN+CLIP Augmentation} We give examples from the VQGAN+CLIP method. Beginning with a sketch of a bus (above) and an airplane (below), we attempt to augment them to the clipart, painting, and real domains. We repeat this procedure for approximately 15\% of images in the sketch domain, picked randomly, before training a classifier on the original data and the augmented data.}
    \label{fig:vqgan_vis} 
\end{figure}

Figure~\ref{fig:vqgan_vis} shows examples of augmented images produced by VQGAN+CLIP on DomainNet. We notice that in some cases the quality of the generated images can be rather poor.

\subsection{Augmentation quality for CUB-Paintings, Colored MNIST, and Waterbirds}
\label{supp:aug_quality}

\begin{figure}[h]
    \centering 
    \includegraphics[width=\textwidth]{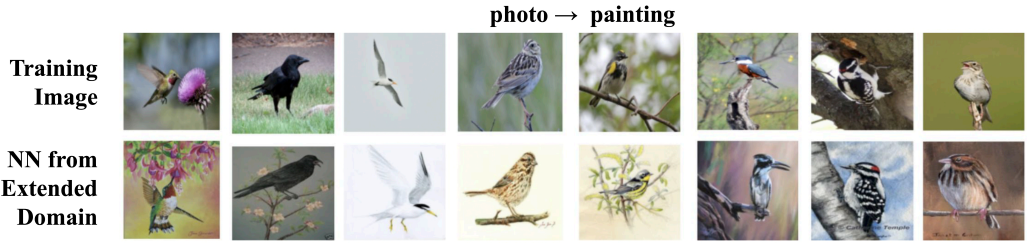}
    \caption{\textbf{Nearest Neighbors for \method{} on CUB-Paintings.} The top row shows training images with the label being the intended domain augmentation for that embedding. The bottom row shows the nearest neighbor of the augmentation in the extended domain. \method{} is able to augment photos to paintings while retaining class-specific information.}
    \label{fig:cub_nn}
\end{figure}

\begin{figure}[h]
    \centering
    \includegraphics[width=\textwidth]{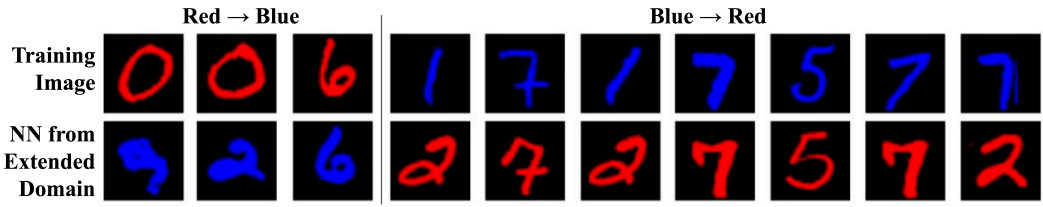}
    \caption{\textbf{Nearest Neighbors for \method{} on Colored MNIST.} The top row shows training images with the the intended domain augmentation for that embedding. The bottom row shows the nearest neighbor of the augmentation in the extended domain. Note that because we do not have a class consistency loss for Colored MNIST due to CLIPS poor performance, the nearest neighbors are often of a different class. }
    \label{fig:mnist_nn}
\end{figure}

Similar to Section~\ref{sec:aug_analysis} of the main paper, where we assess the quality of the augmented image embeddings, Figures \ref{fig:cub_nn} and \ref{fig:mnist_nn} show nearest-neighbor visualizations for the CUB-Paintings and Colored MNIST datasets. Table \ref{tab:aug_quality_overflow} shows the domain alignment and class consistency scores for CUB-Paintings, Colored MNIST, and Waterbirds. We notice that the domain is well-aligned when applying \method{} but the class consistency appears to be lower than on DomainNet. We hypothesize that this is due to the fact that CLIP zero-shot has rather weak performance on CUB-Paintings and MNIST, limiting the capabilities of the class consistency loss. In the future we hope that a class consistency loss which is not dependent on the CLIP zero-shot performance on the task will result in higher class consistency.   


\begin{table}[h]
\small
    \centering
  \begin{tabular}{lrrrrrrr}
    \toprule
     & \multicolumn{2}{c}{CUB-Paintings} & \multicolumn{2}{c}{ColoredMNIST}& \multicolumn{2}{c}{Waterbirds} \\
    \cmidrule(r){2-3} \cmidrule(r){4-5} \cmidrule(r){6-7} 
    Method     & DA & CC & DA & CC & DA & CC \\
    \midrule
    CLIP LP & 99.88$\pm$0.13\% & 73.26$\pm$0.85\% & 98.52$\pm$0.51\% & 96.20$\pm$0.76\% & 91.98$\pm$0.84\% & 95.66$\pm$0.26\% \\
    \method{} & 99.64$\pm$0.27\% & 54.84$\pm$2.69\% & 73.32$\pm$1.48\% & 55.08$\pm$1.79\% & 90.64$\pm$1.00\% & 95.28$\pm$0.66\% \\
    \bottomrule
  \end{tabular}
  \caption{\textbf{Augmentation Quality for CUB-Paintings, Colored MNIST, and Waterbirds.} The domain alignment(DA) and class consistency(CC) scores over 1000 randomly sampled training embeddings and their nearest neighbor in the test set. Note that for Colored MNIST, while the domain is somewhat well aligned, the class of digit often changes, likely due to the fact that we set $\alpha = 1$. }
  \label{tab:aug_quality_overflow}
\end{table}

\clearpage
\subsection{Nearest neighbor visualizations ablating different losses}
\label{supp: nn_ablate_loss}
\begin{figure}[h]
    \centering
    \includegraphics[width=\textwidth]{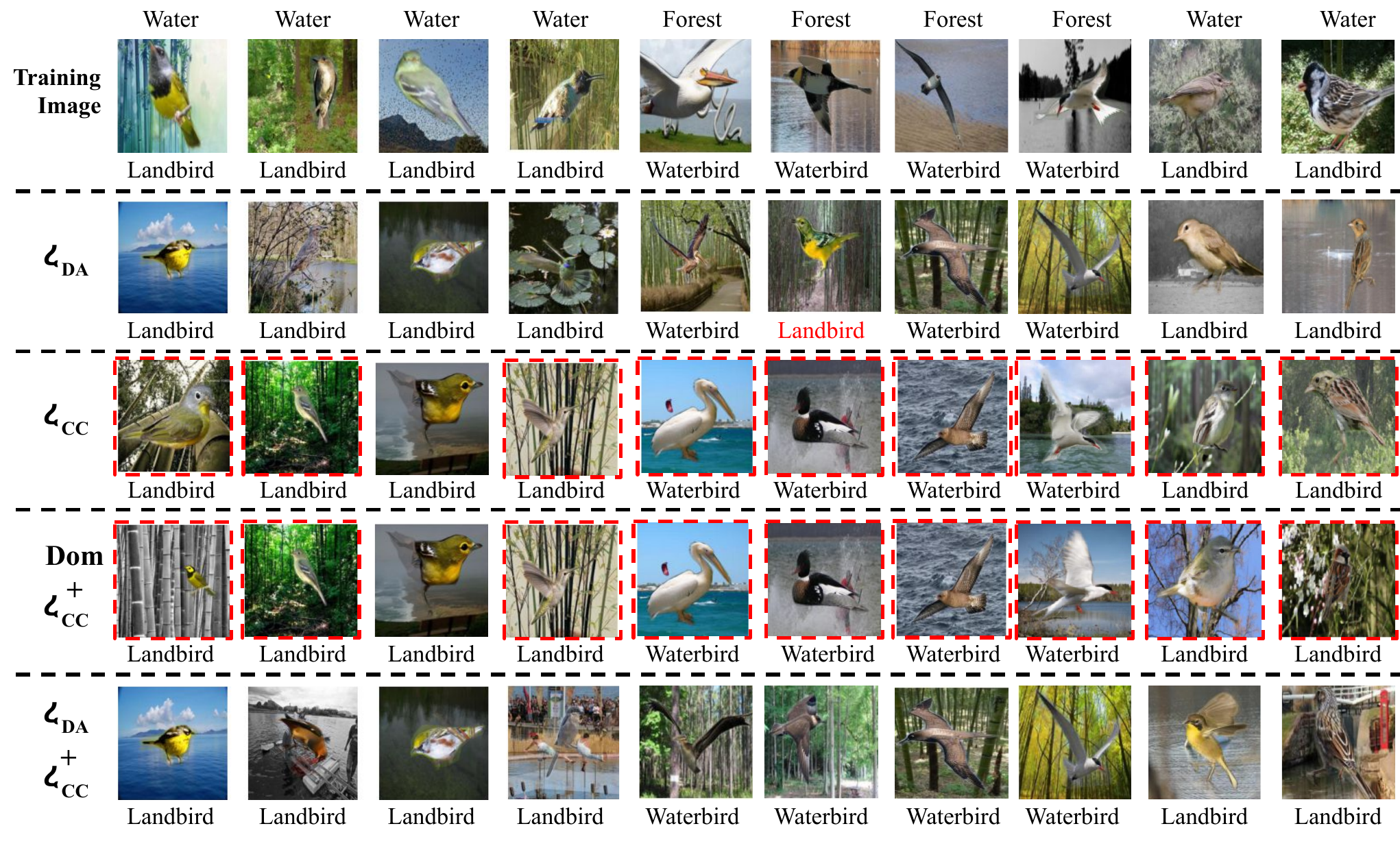}
    \caption{\textbf{Nearest Neighbors for \method{} when ablating the loss on Waterbirds.} The top row shows training images with the label on top being the intended domain augmentation for the image's embedding, and the label on the bottom being the class of the image. The following rows correspond to choices of loss functions as in Section~\ref{sec:loss_ablation} and show the nearest neighbor of the embedding in the test set. The images with red outlines fail to augment to the intended domains, and the ones with red labels are augmented to a different class. Frequently the domain alignment loss is able to augment the embedding into the intended domain although sometimes it does not retain class-specific information. The class consistency loss (both generic and domain specific) retains class-specific information but often fails to augment the embedding to the intended domain.}
    \label{fig:waterbirds_nn_da}
\end{figure}

\begin{figure}[h]
    \centering
    \includegraphics[width=\textwidth]{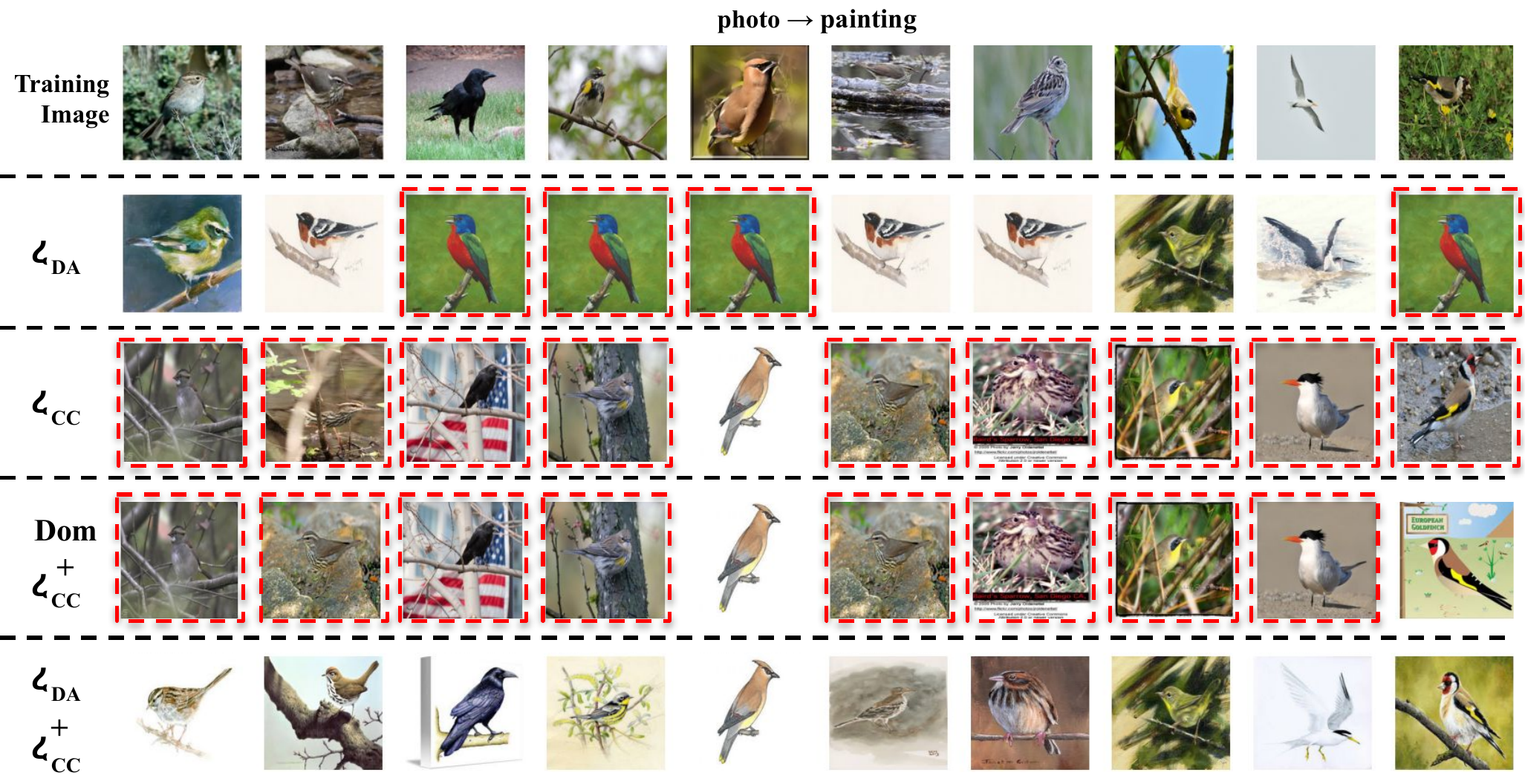}
    \caption{\textbf{Nearest Neighbors for \method{} when ablating the loss on CUB-Paintings.} The images with red outlines fail to augment to the intended domains. While the domain alignment loss is usually able to augment the image into the intended domain, it does not retain class-specific information and often augments to similar features. The class consistency loss (both generic and domain specific) retains class-specific information but fails to convert the image to the intended domain.}
    \label{fig:cub_nn_da}
\end{figure}

Figures~\ref{fig:waterbirds_nn_da} and~\ref{fig:cub_nn_da} compare the nearest neighbors obtained with our full approach and those obtained with the ablations of \method{} for Waterbirds and CUB-Paintings, as defined in Section~\ref{sec:loss_ablation} of the main paper. The images with red outlines fail to augment to the intended domains, and the ones with red labels are augmented to a different class. We see how the domain alignment loss leads to consistent domains (but shifting appearances), while the class consistency losses lead to more similar looking birds (but not necessarily in the right domain). Combining both losses leads to the best accurate nearest neighbors in terms of both, domain alignment and class consistency. Tables~\ref{tab:loss_ablation_Waterbirds} and~\ref{tab:loss_ablation_CUB} report accuracies, domain alignment and class consitency scores from these experiments.


\begin{table}[h]
\small
    \centering
  \begin{tabular}{llllll}
    \toprule
    Method     & ID Acc & OOD Acc & Extended & DA score & CC score \\
    \midrule
    CLIP LP     & {97.77$\pm$0.65}\% & 62.31$\pm$3.79\% & 80.04$\pm$1.59\% & 91.98$\pm$0.84\% & 95.66$\pm$9.26\% \\
    \midrule
    $\mathcal{L}_{\text{DA}}$ & 98.70$\pm$0.41\% & 59.29$\pm$9.35\% & 79.00$\pm$4.47\% & 86.00$\pm$1.78\% & 94.90$\pm$0.86\%\\
    $\mathcal{L}_{\text{CC}}$ & 98.06$\pm$0.72\% & 59.37$\pm$7.28\% & 78.71$\pm$3.29\% & 50.46$\pm$1.93\% & 96.30$\pm$0.19\% \\
    Domain specific $\mathcal{L}_{\text{CC}}$ & 98.18$\pm$0.83\% & 65.46$\pm$1.32\% & 81.82$\pm$0.78\% & 52.04$\pm$1.15\% & 96.20$\pm$0.39\% \\
    $\mathcal{L}_{\text{DA}}$ + $\mathcal{L}_{\text{CC}}$ & 98.09$\pm$ 0.10\% & 71.80$\pm$0.38\% & 84.95$\pm$0.17\% & 90.64$\pm$1.00\% & 95.28$\pm$0.66\%\\
    \bottomrule
  \end{tabular}
  \centering
  \caption{\textbf{Effect of the Loss Functions on Waterbirds}. We report the results of training with just the domain alignment loss, the class consistency loss, a domain-specific class consistency loss, and the domain alignment + class consistency loss, on Waterbirds, corresponding to Figure~\ref{fig:loss_ablation}. The DA loss leads to a high DA score but low accuracy. The CC loss leads to a low DA score and does not improve the OOD accuracy; the domain-specific CC variant brings negligible gains. Our final design (DA+CC losses) works best.}
  \label{tab:loss_ablation_Waterbirds}
\end{table}

\begin{table}[h]
\small
    \centering
  \begin{tabular}{llllll}
    \toprule
    Method     & ID Acc & OOD Acc & Extended & DA score & CC score \\
    \midrule
    CLIP LP     & 95.03$\pm$0.07\% & 93.75$\pm$0.02\%  & 94.39$\pm$0.04\% & 91.42$\pm$0.47\% & 73.32$\pm$1.35\% \\
    \midrule
    $\mathcal{L}_{\text{DA}}$ & 83.87$\pm$ 0.15\% & 53.74$\pm$0.76\% & 68.80$\pm$0.41\% & 37.48$\pm$2.03\% & 89.36$\pm$1.60\%\\
    $\mathcal{L}_{\text{CC}}$ & 85.88$\pm$0.14\% & 64.59$\pm$0.41\% & 75.24$\pm$0.27\% & 9.5$\pm$0.95\% & 57.46$\pm$3.21\% \\
    Domain specific $\mathcal{L}_{\text{CC}}$ & 85.87$\pm$0.05\% & 65.29$\pm$0.19\% & 75.58$\pm$0.12\% & 54.68$\pm$0.54\% & 56.56$\pm$1.13\% \\
    $\mathcal{L}_{\text{DA}}$ + $\mathcal{L}_{\text{CC}}$ & 86.14$\pm$0.29\% & 66.38$\pm$0.25\% & 76.16$\pm$0.23\% & 99.64$\pm$0.27\% & 54.84$\pm$2.69
    \%\\
    \bottomrule
  \end{tabular}
  \centering
  \caption{\textbf{Effect of the Loss Functions on CUB-Paintings}. We report the results of training with just the domain alignment loss, the class consistency loss, a domain-specific class consistency loss, and the domain alignment + class consistency loss, on CUB-Paintings. The DA loss leads to high DA score but low accuracy. The CC loss leads to a low DA score and does not improve the OOD accuracy; the domain-specific CC variant brings negligible gains. Our final design (DA+CC losses) works the best.}
  \label{tab:loss_ablation_CUB}
\end{table}



\subsection{Effect of domain descriptions.}
\label{supp:robustness wrt prompts-appendix}
A consideration when using language to describe domains is that there are different ways to say the same thing. As such we evaluate the robustness of \method{} when using different wordings to describe the same domain shift. Results (obtained on the Waterbirds dataset) are shown in Table~\ref{tab:prompt_ablation}. 
The first two rows, which have similar prompts, have also similar results. In contrast, the bottom two rows shows the results when a target prompt does not match the extended domain. The OOD and Extended domain results are worse with this target prompt. 

\begin{table*}[h]
\scriptsize
\begin{center}
\begin{tabular}{c|c|c|c|c}
\toprule
$\ttrain$ & $\tnew$ & ID & OOD & Extended \\
\hline
“a photo of a \{\} on forest.” & “a photo of a \{\} on water.” & 98.09$\pm$0.10\% & 71.80$\pm$0.38\% & 84.95$\pm$0.17\% \\
“there is a picture of a \{\} on forest.” & “there is a picture of a \{\} on water.” & 98.03$\pm$0.04\% & 72.43$\pm$0.56\% & 85.23$\pm$0.26\% \\
“a photo of a \{\} at disco.” & “a photo of a \{\} at Supreme Court.” & 97.51$\pm$0.12\% & 67.80$\pm$0.98\% & 82.66$\pm$0.45\% \\
“\{\} threaten old seminar.” & “\{\} thank stable student.” & 98.66$\pm$0.08\% & 56.21$\pm$2.39\% & 77.43$\pm$1.16\% \\
\bottomrule
\end{tabular}
\end{center}
\caption{\textbf{Prompt Ablation.} On the Waterbirds dataset, we experiment with similar descriptions of the training and unseen test domain (top two rows) as well as two nonsense prompts (bottom two rows). As show, \method{} produces similar results when given prompts with the same meaning, and obtains comparable or worse results to CLIP LP when given nonsense prompts.}
  \label{tab:prompt_ablation}
\end{table*}

\subsection{Effect of vision and language model}
\label{supp:clip_ablation}
We also experiment over several different CLIP models. Table~\ref{tab:model_ablation} displays how \method{} consistently obtains high accuracy across model sizes, architectures, and pretraining datasets.

\begin{table}[h]
\small
\centering
\begin{tabular}{c|c|c|c}
\toprule
Method     & RN50 (OpenAI) & ViT-L14 (OpenAI) & ViT-H14 (LAION-2B) \\
\midrule
CLIP ZS (G) & 47.96\% & 70.88\% & 64.96\% \\
CLIP ZS (A) & 54.44\% & 78.68\% & 81.64\% \\
\midrule
CLIP LP     & 56.95$\pm$0.06\% & 69.50$\pm$0.25\%  & 86.50$\pm$0.32\% \\
WiSE-FT     & 60.76$\pm$0.31\% & 78.52$\pm$0.23\%  & 92.14$\pm$0.19\% \\
\method{}     & \textbf{69.73$\pm$0.14\%} & \textbf{87.88$\pm$0.13\%} & \textbf{97.12$\pm$0.05\%} \\
\bottomrule
\end{tabular}
\caption{Extended  Accuracy on \textbf{Colored MNIST} using models of ResNet50 and ViT\text{-}L\/14 from OpenAI and ViT\text{-}H\/14 pretained on LAION\text{-}2B from OpenCLIP \citep{ilharco_gabriel_2021_5143773}.}
    \label{tab:model_ablation}
\end{table}

\section{Failure case: MNIST-to-SVHN adaptation}
\label{supp:mnist-svhn-appendix}

We show results of adapting MNIST to SVHN in Table~\ref{tab:mnist_svhn}, another domain adaptation benchmark. While \method{} outperforms the fine-tuning baselines, it has poor performance when compared to zero-shot. We suspect this is because CLIP is bad at differentiating an MNIST digit from an SVHN digit, achieving only around 20\% accuracy at classifying the digit domain.

\begin{table}[h]
\small
\centering

\begin{tabular}{c|c|c|c}
\toprule
Method &	MNIST (ID) &	SVHN (OOD) &	Extended \\
\midrule
CLIP ZS	& 80.60\% & 	41.44\% & 	61.02\% \\
CLIP ZS (A)	& 97.02\% & 	\textbf{57.3\%} & 	\textbf{77.16\%}\\
\midrule
CLIP LP	& \textbf{97.2$\pm$0.00\%}	&49.63 $\pm$0.10\%&	73.36$\pm$0.10\% \\
WiSE & \textbf{97.2$\pm$0.00\%}	&49.63 $\pm$0.10\%&	73.36$\pm$0.10\% \\
\method{} &	\textbf{97.2$\pm$0.00\%} &	51.99$\pm$0.12\%&	74.60$\pm$0.06\%  \\
\bottomrule
\end{tabular}
\caption{In-domain (ID), out-of-domain (OOD) and extended domain accuracy on MNIST to SVHN.}
  \label{tab:mnist_svhn}
\end{table}

\section{Effect of $\alpha$}
While our loss ablation (Sec ~\ref{sec:loss_ablation}) is exploring $\alpha = 0$ (only $\mathcal{L}_{\text{CC}}$) and $\alpha = 1$ (only $\mathcal{L}_{\text{DA}}$), we explore other values of $\alpha$ on CUB Paintings and Waterbirds in Table ~\ref{tab:cub_alpha_ablation} and Table ~\ref{tab:waterbirds_alpha_ablation}. Note that the domain alignment (DA) and class consistency (CC) scores are the proportion of nearest neighbors of the augmented embeddings that match the desired (target) domain and class.

As shown in Sec ~\ref{sec:loss_ablation}, when alpha = 0, the class consistency is high and the domain alignment is low, while the opposite is seen when alpha = 1. However, we see that we can often achieve a better domain alignment score when alpha is between 0 and 1. We believe that this is because the augmented embeddings drift too far from the unaugmented embeddings without the CLIP supervision that the class consistency score provides. 

\begin{table}[h]
\small
    \centering
  \begin{tabular}{llllll}
    \toprule
    $\alpha$     & ID Acc & OOD Acc & Extended & DA score & CC score \\
    \midrule
    0 & 97.72$\pm$0.23\% & 66.97$\pm$1.48\% & 82.34$\pm$0.66\% & 32.12$\pm$1.33\% & 95.46$\pm$0.70\%\\
    0.25 & 99.96$\pm$0.10\% & 72.95$\pm$0.40\% & 85.45$\pm$0.20\% & 67.22$\pm$2.95\% & 94.90$\pm$0.75\% \\
    0.5 & 98.03$\pm$0.06\% & 72.23$\pm$0.47\% & 85.13$\pm$0.21\% & 89.60$\pm$3.26\% & 92.86$\pm$1.52\% \\
    0.75 & 98.13$\pm$ 0.03\% & 71.65$\pm$0.40\% & 84.89$\pm$0.18\% & 95.22$\pm$0.64\% & 89.50$\pm$1.21\%\\
    1.0 & 98.69$\pm$ 0.35\% & 63.77$\pm$10.60\% & 81.23$\pm$5.13\% & 88.10$\pm$1.83\% & 94.30$\pm$1.23\%\\
    \bottomrule
  \end{tabular}
  \centering
  \caption{\textbf{Effect of the $\alpha$ for \method{} on Waterbirds}.}
  \label{tab:waterbirds_alpha_ablation}
\end{table}

\begin{table}[h]
\small
    \centering
  \begin{tabular}{llllll}
    \toprule
    $\alpha$     & ID Acc & OOD Acc & Extended & DA score & CC score \\
    \midrule
    0 & 85.88$\pm$0.14\% & 64.59$\pm$0.41\% & 75.24$\pm$0.27\% & 9.50$\pm$0.95\% & 57.46$\pm$3.21\%\\
    0.25 & 85.89$\pm$0.17\% & 65.04$\pm$0.73\% & 75.47$\pm$0.45\% & 89.16$\pm$3.21\% & 51.28$\pm$5.93\% \\
    0.5 & 86.14$\pm$0.29\% & 66.18$\pm$0.25\% & 76.16$\pm$0.23\% & 99.64$\pm$0.27\% & 54.84$\pm$2.69\% \\
    0.75 & 85.63$\pm$ 0.09\% & 62.93$\pm$0.31\% & 74.28$\pm$0.18\% & 94.20$\pm$1.00\% & 44.56$\pm$2.27\%\\
    1.0 & 83.87$\pm$ 0.15\% & 53.74$\pm$0.76\% & 68.80$\pm$0.41\% & 89.36$\pm$1.60\% & 37.48$\pm$2.03\%\\
    \bottomrule
  \end{tabular}
  \centering
  \caption{\textbf{Effect of the $\alpha$ for \method{} on CUB Paintings}.}
  \label{tab:cub_alpha_ablation}
\end{table}

\section{When to use \method{} over CLIP ZS}

We analyze when to use LADS over CLIP ZS. In general, CLIP ZS should be used over LADS when linear probing the CLIP embeddings outperforms ZS on the majority of classes from the source domain.

For example, in Table\ref{tab:officehome} we have the results for 2 splits of OfficeHome~\citep{officehome}: source domain of clipart and source domain of product, with the test domain being all 4 domains (art, clipart, product, real world). For both splits we see the same phenomenon: since CLIP ZS outperforms CLIP LP on 37/65 classes on the source domain, the overall OOD accuracy of CLIP ZS is higher than LADS. However, if we take the 28/65 classes where CLIP LP outperforms CLIP ZS, then we see that LADS OOD accuracy beats ZS and CLIP LP. 
We have added a section in the Appendix explaining this phenomenon.

\begin{table}[h]
\small
    \centering
    \begin{tabular}{cc|cc|cc}
    \toprule
    \multicolumn{2}{c|}{} & \multicolumn{2}{c|}{$\dtrain= Clipart$} & \multicolumn{2}{c}{$\dtrain=Product$} \\
    \cmidrule{3-4} \cmidrule{5-6} 
    Classes Subset & Method & ID & OOD & ID & OOD\\
    \midrule[2pt]
    All (65/65) & CLIP ZS (G) & 72.07\% & 90.19\% & 94.61\% & \textbf{86.13\%} \\
    All (65/65) & CLIP ZS (A) & 74.75\% & \textbf{91.34\%} & 94.49\% & 84.73\% \\
    \midrule
    All (65/65) & CLIP LP & \textbf{82.66$\pm$0.16\%} & 85.95$\pm$0.65\% & 95.46$\pm$0.03 & 78.53$\pm$0.40 \\
    All (65/65) & \method{} & \textbf{82.55$\pm$0.28\%} & 87.97$\pm$0.50\% & \textbf{96.21$\pm$0.20\%} & 81.20$\pm$0.22\% \\
    \midrule[2pt]
    ZS Win (37/65) & CLIP ZS (G) & 79.64\% & 89.70\% & 97.19\% & \textbf{88.55\%} \\
    ZS Win (37/65) & CLIP ZS (A) & \textbf{80.95\%} & \textbf{91.32\%} & \textbf{97.42\%} & 87.86\% \\
    \midrule
    ZS Win (37/65) & CLIP LP & 79.44$\pm$0.17\% & 81.92$\pm$0.69\% & 95.79$\pm$0.40\% & 75.17$\pm$0.64\% \\
    ZS Win (37/65) & \method{} & 79.15$\pm$0.61\% & 84.68$\pm$ 0.77\% & 96.16$\pm$0.25\% & 78.79$\pm$0.37\% \\
    \midrule[2pt]
    LP Win (28/65) & CLIP ZS (G) & 62.08\% & 90.84\% & 90.72\% & 82.49\% \\
    LP Win (28/65) & CLIP ZS (A) & 66.56\% & 91.36\% & 90.09\% & 80.04\% \\
    \midrule
    LP Win (28/65) & CLIP LP     & \textbf{86.91$\pm$0.41\%} & 91.28$\pm$0.06\% & 94.97$\pm$0.67\% & 83.56$\pm$0.09\% \\
    LP Win (28/65)
    & \method{}     & \textbf{87.20$\pm$0.28\%} & \textbf{92.31$\pm$ 0.16\%} & \textbf{96.29$\pm$0.32\%} & \textbf{84.82$\pm$0.13\%} \\
    \bottomrule
  \end{tabular}
  \caption{In-domain (ID), out-of-domain (OOD) and extended domain accuracy on \textbf{OfficeHome} with a source domain of clipart and product and a test set of all 4 domains (art, clipart, product, real world) over different subsets of classes. On classes where ZS outperforms CLIP LP (37/65 classes) on the validation set, zero-shot outperforms \method{} on OOD and extended accuracy. The opposite phenomenon is seen on the subset of classes where CLIP LP matches or beats ZS on the validation set.} 
  \label{tab:officehome}
\end{table}


\end{document}